\documentclass[graybox]{IFCS2024}
\usepackage{type1cm}        
\usepackage{makeidx}        
\usepackage{graphicx}      
\usepackage{multicol}       
\usepackage[bottom]{footmisc}
\usepackage{newtxtext}       
\usepackage[varvw]{newtxmath}   
\usepackage{algorithm}
\usepackage{algpseudocode}  

\begin{document}

\title*{Riemannian Principal Component Analysis}

\titlerunning{Riemannian Principal Component Analysis}

\author{Oldemar Rodríguez}

\institute{Oldemar Rodríguez \at School of Mathematics, CIMPA, University of Costa Rica, San José, Costa Rica,
 \email{oldemar.rodriguez@ucr.ac.cr}}

\maketitle

\abstract*{This paper proposes an innovative extension of Principal Component Analysis (PCA) that transcends the traditional assumption of data lying in Euclidean space, enabling its application to data on Riemannian manifolds. The primary challenge addressed is the lack of vector space operations on such manifolds. Fletcher et al., in their work {\em Principal Geodesic Analysis for the Study of Nonlinear Statistics of Shape}, proposed Principal Geodesic Analysis (PGA) as a geometric approach to analyze data on Riemannian manifolds, particularly effective for structured datasets like medical images, where the manifold's intrinsic structure is apparent. However, PGA's applicability is limited when dealing with general datasets that lack an implicit local distance notion.  In this work, we introduce a generalized framework, termed {\em Riemannian Principal Component Analysis (R-PCA)}, to extend PGA for any data endowed with a local distance structure. Specifically, we adapt the PCA methodology to Riemannian manifolds by equipping data tables with local metrics, enabling the incorporation of manifold geometry. This framework provides a unified approach for dimensionality reduction and statistical analysis directly on manifolds, opening new possibilities for datasets with region-specific or part-specific distance notions, ensuring respect for their intrinsic geometric properties.}

\abstract{This paper proposes an innovative extension of Principal Component Analysis (PCA) that transcends the traditional assumption of data lying in Euclidean space, enabling its application to data on Riemannian manifolds. The primary challenge addressed is the lack of vector space operations on such manifolds. Fletcher et al., in their work {\em Principal Geodesic Analysis for the Study of Nonlinear Statistics of Shape}, proposed Principal Geodesic Analysis (PGA) as a geometric approach to analyze data on Riemannian manifolds, particularly effective for structured datasets like medical images, where the manifold's intrinsic structure is apparent. However, PGA's applicability is limited when dealing with general datasets that lack an implicit local distance notion.  In this work, we introduce a generalized framework, termed {\em Riemannian Principal Component Analysis (R-PCA)}, to extend PGA for any data endowed with a local distance structure. Specifically, we adapt the PCA methodology to Riemannian manifolds by equipping data tables with local metrics, enabling the incorporation of manifold geometry. This framework provides a unified approach for dimensionality reduction and statistical analysis directly on manifolds, opening new possibilities for datasets with region-specific or part-specific distance notions, ensuring respect for their intrinsic geometric properties.}

.

\section{Introduction}
\label{sec:int}

Each point in a data table can be imagined as a star or planet in the universe, especially when dealing with big data issues. In the universe, due to the infinitely different sizes of constellations, there are vastly different perceptions of distances between celestial bodies. For example, two constellations or galaxies that appear to be the same size from a distance (from Earth, for example) could be infinitely different, and one could even fit inside the other in a very small portion or empty space within it. For this reason, especially in problems involving Big Data, {\em thinking that the data is in Euclidean space is just as wrong as thinking that the earth is flat}.

Similarly, in data, there are local notions of distance corresponding to different regions of the data, and this should be considered when calculating indices or statistical models, as Principal Component Analysis. To address this, we propose considering that the data exists within a Riemannian manifold, where these local notions of distance can be effectively taken into account.

In  \cite{flet} {\em Fletcher et al.}, the authors had proposed the Principal Geodesic Analysis on Riemannian manifolds through the use of geometry. This concept works particularly well when analyzing data derived from images, such as medical images, where the intrinsic Riemannian manifold structure is evident. However, this idea is not readily applicable to general data where there are no implicit notions of local distance. The {\em Riemannian Principal Component Analysis}  that we propose go beyond of what was mentioned in the previous paragraph. The core concept is to impart a Riemannian manifold structure to any given set of data. This approach enables the assignment of local notions of distance to the data, thereby enhancing our ability to capture the internal structure of the data. This, in turn, leads to a significant improvement in the results of various statistical analyses as well as their interpretability.

\begin{figure}[h]
\sidecaption
\includegraphics[scale=.24]{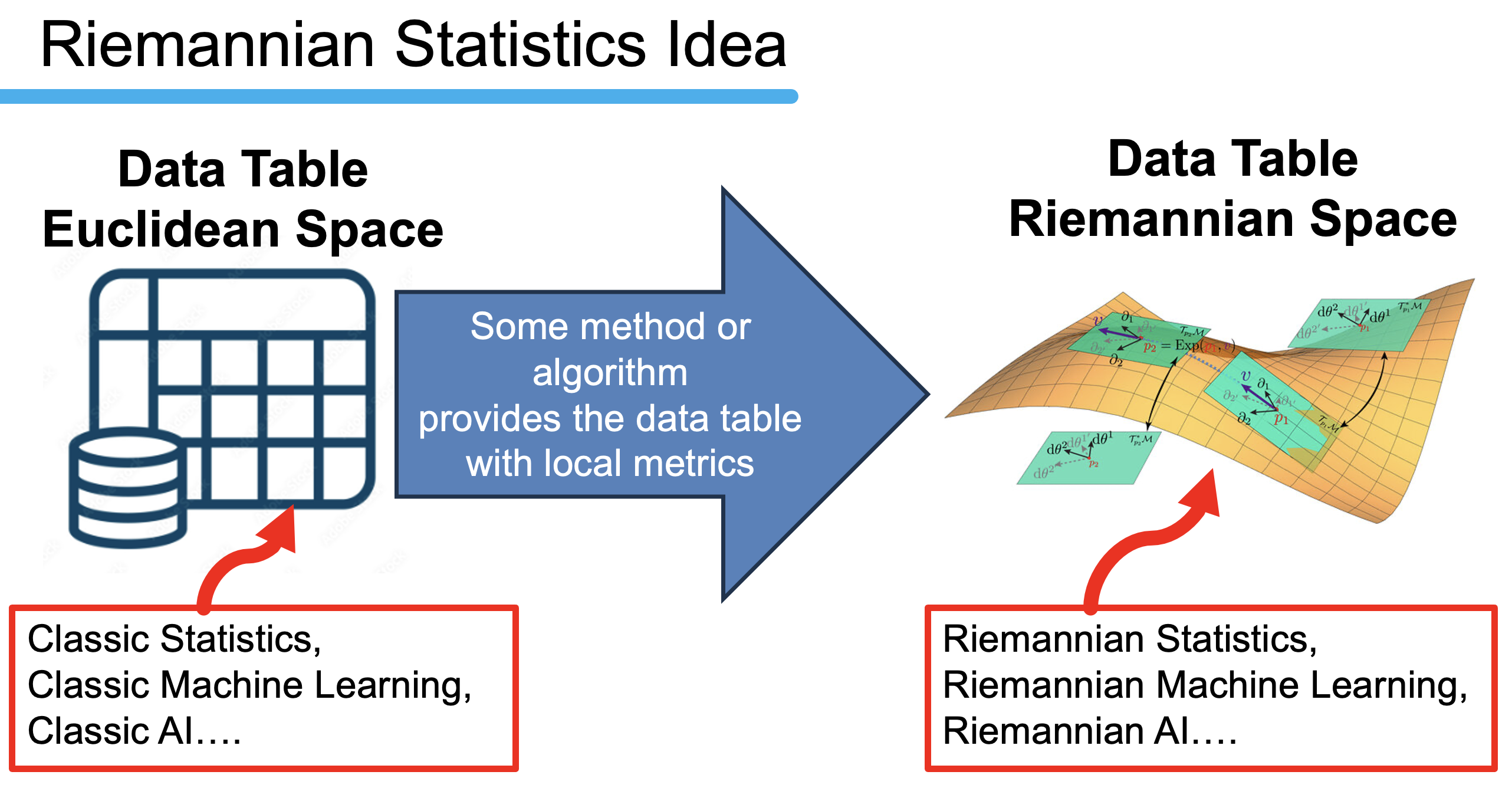}
\caption{Transforming an Euclidean into a Riemannian Space.}
\label{fig:PCA1}       
\end{figure}

The diagram in Figure \ref{fig:PCA1} illustrates the transformation of data analysis methodologies from a Euclidean framework to a Riemannian geometric framework. It begins with a data table residing in Euclidean space, where classical techniques such as statistics, machine learning, and artificial intelligence are traditionally applied, relying on global distance metrics and vector space operations. By applying a specific method or algorithm, the data is endowed with local metrics that capture intrinsic geometric properties, enabling a transition to a Riemannian space. In this Riemannian space, data analysis leverages curvature, local distances, and local structures, transforming classical methodologies into Riemannian Statistics, Riemannian Machine Learning, and Riemannian AI. This paradigm shift allows for more accurate and meaningful analyses of data that naturally resides on curved manifolds, such as shapes or other structured datasets, by respecting their underlying geometric nature. 

In this work, we propose to transform the Euclidean space into a Riemannian manifold using the UMAP (Uniform Manifold Approximation and Projection) algorithm, see McInnes et al. \cite{mcinnes}. As described in the paper {\em UMAP: Uniform Manifold Approximation and Projection for Dimension Reduction}, the transformation is based on the construction of local metrics derived from a graph representation of the data. Specifically, UMAP methodology, detailed in Section 3.1 Graph Construction of the referenced paper, $k$-nearest neighbor graph where local distances between data points are computed and normalized. These distances are then used to define local Riemannian metrics, effectively approximating a Riemannian structure for the data manifold. By leveraging the graph local connectivity properties and its edge weights, UMAP provides an efficient and scalable way to construct Riemannian metrics, enabling the analysis of complex datasets in their intrinsic geometric space. This transformation allows for the application of Riemannian statistics and machine learning methods, capturing both the local and global structures of the data.

UMAP is a novel technique for manifold learning and dimension reduction. Utilizing simplicial complexes, Čech complexes, and the Nerve theorem, UMAP gains additional benefits from this Riemannian metric-based approach. It generates a local metric space associated with each point, allowing for meaningful distance measurements. Consequently, the algorithm can assign weights to edges in a graph (simplicial complex), signifying the local metric-based separation between the original points. So the idea that we proposed in this paper is to use the local notions of distance that the UMAP algorithm generates in any data table to provide it with local distance. In this way, the data table can be conceptualized as a Riemannian manifold, incorporating these local distance.

UMAP, as a successor to $t$-SNE method, inherits a controversy associated with the $t$-SNE method. The challenge with $t$-SNE lies in its inability to preserve distances and density effectively. It only partially maintains the concept of {\em nearest-neighbors}. Though the distinction may seem subtle, it has implications for any clustering algorithm based on density or distance. This issue is somewhat  controversial, and should be approached with caution.  A comprehensive discussion on this topic can be found at \url{https://umap-learn.readthedocs.io/en/latest/clustering.html}.

Despite these concerns, there are still valid reasons to utilize UMAP as a preprocessing step for clustering. As highlighted in the discussion, when applied to real high-dimensional datasets such as {\tt MNIST} data \cite{deng} or {\tt cell RNA-seq} data \cite{karthik}, and with appropriate parameterization, both $t$-SNE and UMAP yield significantly better clustering results than other algorithms. Regardless, for Riemannian Principal Component Analysis, the crucial aspect is that UMAP maintains the concept of {\em nearest-neighbors} in the low-dimensional representation of the dataset. This is very important as it provides the data table with local distance notions, enhancing the utility of the UMAP algorithm in this context.

\section{Providing to a classical data table with a Riemannian manifold structure}
\label{sec:mani}

UMAP method was designed to improve the main limitations of the $t$-SNE method.  $t$-SNE means $t$-distributed Stochastic Neighbor Embedding and it was proposed by Laurens van der Maaten, see all the detail of this method in \cite{maaten}. UMAP algorithm is competitive with $t$-SNE for visualization quality and it improves $t$-SNE limitations. UMAP (Uniform Manifold Aproximation and Projection) is an algorithm for dimension reduction based on algebraic topology, topological data analysis and Riemannian geometry. It was proposed by the Mathematician Leland McInnes in \cite{mcinnes}. UMAP works in a similar way to $t$-SNE, it finds distances in a space with many variables and then tries to reproduce these distances in a low-dimensional space. But UMAP does it very differently because more than distances it tries to reproduce the topology, not necessarily the geometry. UMAP assumes that data is distributed along a Riemannian manifold. A manifold is a uniform $n$-dimensional geometric shape in which, for each point of this manifold, there is a neighborhood around that point that looks like a flat two-dimensional plane. Riemannian manifolds admit local notions of distances, area and angles. To explain the UMAP method we need to define the notion of $k$-simplex and simplicial complexes.  

Let $\{x_0, \ldots, x_k\}$ be points in $\mathbb{R}^n$. We will assume that these points satisfy the condition that the set of vectors in $\mathbb{R}^n$ represented by the differences with respect to $x_0$, that is $\{x_1 - x_0, x_2 - x_0, \ldots, x_k - x_0\}$
are linearly independent. 

\begin{definition}
The $k$-simplex generated by the points $\{x_0, \ldots, x_k\}$ is the set of all points
$ z = \sum_{i=0}^{k} a_i x_i,$
where
$ \sum_{i=0}^{k} a_i = 1. $
For a given $z$, we refer to $a_i$ as the $i$-th barycentric coordinate.
\end{definition}

Simplicial complexes are generalizations of graphs. A simplicial complex $S$ in $\mathbb{R}^n$ is a set of simplices such that every face of a simplex in $S$ is also a simplex in $S$. The intersection of two simplices in $S$ is a face of each of them.
Given data set presented as a finite metric space, we need to produce a simplicial complex such that the algebraic invariants of the simplicial complex reflect the shape of the data. To do that, we need to make the connection between clustering and components precise, via single-linkage clustering, which works as follows.

\begin{figure}[h]
\sidecaption
\includegraphics[scale=.30]{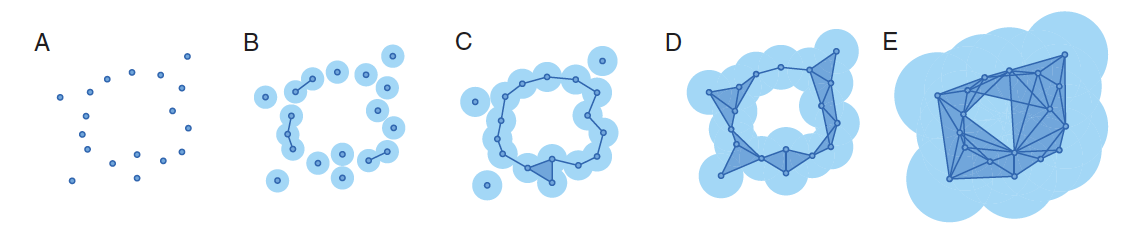}
\caption{As \(\epsilon\) increases, more and more simplices are added to the simplicial complex and topological features emerge. In panels \(C\) and \(D\), a circle can be detected.}
\label{fig:A}       
\end{figure}

\begin{enumerate}
    \item Choose a parameter \(\epsilon\).
    \item Assign two points \(x\) and \(y\) to the same group if they are connected by a path of points (for some \(k\))
    $
    x = x_{0}, x_{1}, x_{2}, \ldots, x_{k-1}, x_{k} = y
    $
    such that each point \(x_{i}\) is at a distance \(\epsilon\) from \(x_{i+1}\). See the Figure \ref{fig:A}.
\end{enumerate}

The Nerve Theorem and its corollary are the fundamental theoretical basis that allows us to go from topological spaces to simplicial complexes and then to data. The Čech complex allows us to demonstrate that there exists a homeomorphism between the union of balls (determined by the parameter \(\epsilon\)) and the nerve and therefore we will have a bijection between the data and the simplicial complexes.

\begin{definition}
The nerve $N(\mathcal{U})$ of a cover $ \mathcal{U} = \left\{U_{i}\right\}$ of topological space  \(X\) is the simplicial complex
with vertices corresponding to the sets \(\left\{U_{i}\right\}\) and a \(k\)-simplex \(\left[j_{0}, j_{1}, \ldots, j_{k}\right]\) when the intersection $
U_{j_{0}} \cap U_{j_{1}} \cap U_{j_{2}} \cap \cdots \cap U_{j_{k}} \neq \emptyset.
$
\end{definition}

\begin{definition}
Let \(X \subset \mathbb{R}^{n}\) be a finite subspace and fix \(\epsilon>0\).
The Čech complex \(C_{\epsilon}\left(X, \partial_{X}\right)\) is the simplicial complex with vertices the points of \(X\), and
a \(k\)-simplex \(\left[v_{0}, v_{1}, \ldots, v_{k}\right]\) when a set of points   \(\left\{v_{0}, v_{1}, \ldots, v_{k}\right\} \subset X\) satisfies
$
\bigcap_{i} B_{\epsilon}\left(v_{i}\right) \neq \emptyset.
$
\end{definition}

\begin{theorem} [Nerve Theorem] Let \(X\) be a topological space.  Let $ \mathcal{U} = \left\{U_{i}\right\}$ be an open cover of \(X\)
such that all non-empty finite intersections
$
U_{j_{1}} \cap U_{j_{2}} \cap \cdots \cap U_{j_{k}}
$
are contractible (homotopy equivalent to a point). Then the nerve (the geometric realization) $N(\mathcal{U})$ is homotopy equivalent to \(X\).
\label{TN}
\end{theorem}

\begin{corollary}
Let \(X \subset \mathbb{R}^{n}\) be a finite subspace and fix \(\epsilon>0\). There exists a homeomorphism:
$
\bigcup_{x \in X} B_{\epsilon}(x) \cong\left|C_{\epsilon}\left(X, \partial_{X}\right)\right|
$
between the union of balls and the nerve $N(\mathcal{U})$ (the geometric realization) of the Čech complex.
\end{corollary}

\begin{figure}[h]
\sidecaption
\includegraphics[scale=.28]{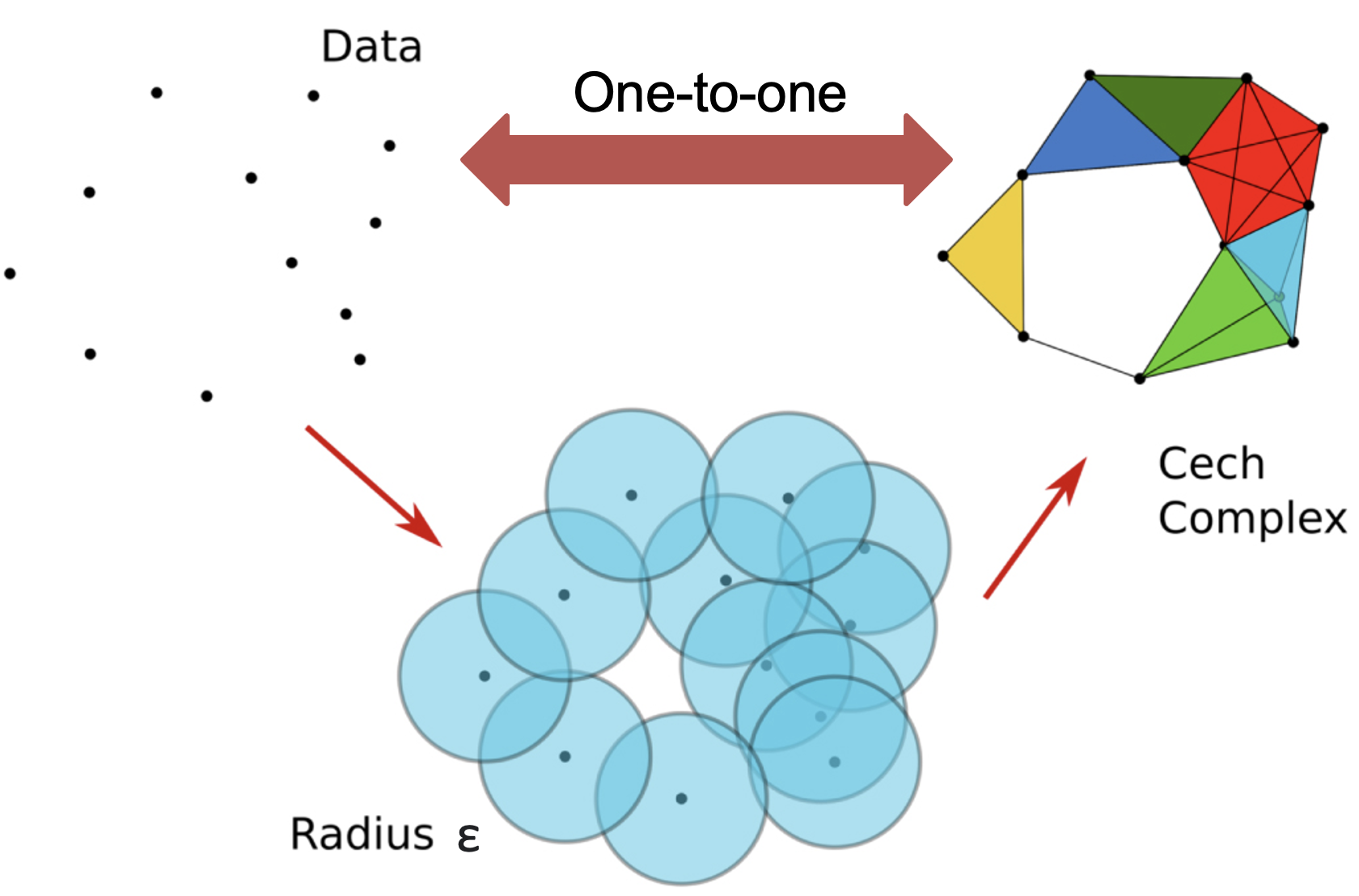}
\caption{Relation one-to-one between data and Čech complex.}
\label{fig:B}       
\end{figure}

The above guarantees that there exists a homeomorphism between the union of balls and the nerve, so, there is relation one-to-one (bijection) between data and Čech complex, as it is illustrated in the Figure \ref{fig:B}.

To apply these ideas, UMAP choose a radius from each point, connecting points when those radii overlap, then we can create a simplicial complex using 0, 1, and 2 simplexes as points, lines, and triangles. Choosing this radius is critical, too small choice will lead to small, isolated clusters, while too large choice will connect everything together. UMAP overcomes this challenge by choosing a radius locally, based on the local distance to each point to the $k$-th nearest neighbor. To do that, Riemannian Geometry is used.

\begin{definition}
Fixed $x$, a {\bf Riemannian metric} is defined by a  scalar products \(\langle\cdot, \cdot\rangle_{x}\) on each tangent space \(T_{x} \mathcal{M}\) at points
\(x\) of the manifold.  For each \(x\),  each such scalar product is a positive definite bilinear map
$ \langle\cdot, \cdot\rangle_{x}: T_{x} \mathcal{M} \times T_{x} \mathcal{M} \rightarrow \mathbb{R}.$
The inner product gives a norm \(\|\cdot\|_{x}: T_{x} \mathcal{M} \rightarrow \mathbb{R}\) by \(\|v\|^{2}=\langle v, v\rangle_{x}\)
\end{definition}

The choice of $k$ determines how locally we wish to estimate the Riemannian metric. A small choice of $k$ means we want a very local interpretation, while, choosing a large $k$ means our estimates will be based on larger regions. {\em This is very important, because it means that the UMAP algorithm provides the data table with local distance notions.} To the graph construction, UMAP algorithm begins by constructing a weighted \(k\)-nearest neighbor graph from the dataset \(X = \{x_1, \ldots, x_n\}\), where a distance metric \(d\) defines the distances between points. The process can be described as follows:

\begin{enumerate}
\item The dataset \(X\) consists of points in a high-dimensional space, and a distance metric \(d\) (e.g., Euclidean distance) is used to compute the similarity between points.

\item The algorithm identifies the \(k\)-nearest neighbors using the distance metric \(d\). This step produces a directed graph where each point has outgoing edges to its \(k\)-nearest neighbors.

\item The local parameters, \(\rho_i\) and \(\sigma_i\), are defined as follows:

\begin{itemize}
    \item \(\rho_i\): Local Connectivity. For each \(x_i\), \(\rho_i\) is the minimum distance to its nearest neighbor that is greater than zero:
    
    \[
    \rho_i = \min\{d(x_i, x_{i_j}) \mid 1 \leq j \leq k, \ d(x_i, x_{i_j}) > 0\}.
    \]
    
    This ensures that \(x_i\) connects to at least one other point with an edge of weight 1.

    \item \(\sigma_i\): Local Scaling. \(\sigma_i\) normalizes the distances of all neighbors, ensuring consistency in the local metric. It is computed by solving:
    
    \[
    \sum_{j=1}^k \exp\left(-\frac{\max(0, d(x_i, x_{i_j}) - \rho_i)}{\sigma_i}\right) = \log_2(k).
    \]
    
\end{itemize}

\item To constructing the weighted directed graph we use \(\rho_i\) and \(\sigma_i\), and the weight of each directed edge \((x_i, x_{i_j})\) is computed as:

\[
w((x_i, x_{i_j})) = \exp\left(-\frac{\max(0, d(x_i, x_{i_j}) - \rho_i)}{\sigma_i}\right).
\]

\item \label{simetric} To symmetrized the graph we convert the directed graph into an undirected graph, the adjacency matrix \(A\) is symmetrized using:

\[
B = A + A^\top - A \circ A^\top,
\]

where \(A\) is the adjacency matrix, \(\circ\) denotes the Hadamard product, and \(\top\) represents the transpose matrix. The resulting symmetric graph represents the unified global manifold structure of the dataset.

\item The symmetrized graph \(G\) forms a \textit{fuzzy simplicial set} that captures both the local and global topological structure of the data. Also, this symmetrized graph \(G\) allows us to define a local similarity as $S_{\textsf{\tiny{UMAP}}}({x_i},{x_j})=B_{ij}$. In the UMAP algorithm, the values in the distance graph represent the normalized weights of the connections between points in the $k$-nearest neighbor ($k$-NN) graph. These values range from 0 to 1, as UMAP normalizes these relationships to compute probabilistic affinities (similarities). A value of 0 indicates no connection between the points in the graph, while a value greater than 0 signifies a connection, with the value representing the strength of the connection. A value close to 1 indicates a strong connection.
\end{enumerate}

To generalize Principal Component Analysis it will be important to understand how UMAP connects points from different neighborhoods because this will have implications on how to project individuals belonging to different Riemannian submanifolds. In UMAP, connections between points that belong to different neighborhoods (i.e., points that the $k$-nearest neighbors ($k$-NN) algorithm did not place in the same neighborhood) are handled during the \textit{graph symmetrization step} explained above in \ref{simetric}. 
If two points \(x_i\) and \(x_j\) may belong to different neighborhoods, for example, \(x_i\) may not consider \(x_j\) one of its \(k\)-nearest neighbors, and vice versa, then it may be no direct edge between \(x_i\) and \(x_j\) in the initial directed graph. To address this, UMAP symmetrizes the directed graph by combining local neighborhood information from both points. 
The symmetrization step combines forward and backward edge weights, creating an undirected edge between points in different neighborhoods. The edge weights are symmetrized as:

\begin{equation} \label{sym}
w_{ij} = w_{ij}^{\rightarrow} + w_{ij}^{\leftarrow} - w_{ij}^{\rightarrow} \cdot w_{ij}^{\leftarrow},
\end{equation}

\noindent where \( w_{ij}^{\rightarrow} \) and \( w_{ij}^{\leftarrow} \) are the forward and backward affinities between \( x_i \) and \( x_j \), computed from their respective local neighborhoods. This ensures points in separate neighborhoods can still be connected if their neighborhoods overlap or have similar affinities, capturing both local and global structures of the data. Also, this ensures that redundant weights from overlapping neighborhoods are not double-counted.

\section{Riemannian Principal Component Analysis for any type of data}
\label{sec:mani}

In the paper \cite{flet}, they aim to generalize Principal Component Analysis (PCA) to Principal Geodesic Analysis (PGA) a generalization of principal component analysis to manifolds using geodesic distances and geodesic submanifolds.

Let be $x_1, \dots, x_n \in \mathbb{R}^p$ with zero mean. Principal component analysis find an orthonormal basis $\{v_1, \dots, v_p\}$ of $\mathbb{R}^p$, which satisfies the recursive relationship:

\begin{align}
v_1 &= \arg \max_{\|v\| = 1} \sum_{i=1}^n (v \cdot x_i)^2, \label{eq:v1} \\
v_s &= \arg \max_{\|v\| = 1} \sum_{i=1}^n \sum_{j=1}^{s-1} (v_j \cdot x_i)^2 + (v \cdot x_i)^2, \quad s = 2, \dots, p. \label{eq:vk}
\end{align}

Then, the subspace $V_s = \operatorname{span}(\{v_1, \dots, v_s\})$ is the $s$-dimensional subspace that maximizes the variance of the data projected onto that subspace. As is well known, the basis $\{v_s\}$ is computed as the set of ordered eigenvectors of the sample covariance matrix of the data.

The lower-dimensional subspaces in PCA are linear subspaces, in \cite{flet} for general manifolds $H$, they extend the concept of a linear subspace to a geodesic submanifold. A geodesic is a curve that is locally the shortest path between points. In this way, a geodesic is the generalization of a straight line. Thus, it is natural to use a geodesic curve as the one-dimensional subspace, the analog of the first principal direction in PCA.

Let $M$ be a manifold, the projection of a point $x \in M$ onto a geodesic submanifold $H$ of $M$ is defined as the point on $H$ that is nearest to $x$ in geodesic distance. So, the projection operator $\pi_H : M \to H$ is:

\begin{equation} \label{proy}
\pi_H(x) = \arg \min_{y \in H} d(x, y)^2.
\end{equation}

Since projection is defined by a minimization, there is no guarantee that the projection of a point exists or that it is unique. However, by restricting to a small enough neighborhood about the mean, the projection is unique for any geodesic submanifold  at the mean and the projection onto a geodesic submanifold can be approximated linearly in the tangent space of $M$,   that is,  in \ref{proy} to ensure that $\pi_H(x)$ is actually within $H$, an approximation of the projection is used,  then they find a sequence of nested geodesic submanifolds that maximize the projected variance of the data.  Finally,  given a dataset \( x_1, x_2, \dots, x_n \in M \),  they compute the intrinsic mean \( \mu \), which minimizes the sum of squared Riemannian distances, \( \mu = \arg \min_{x \in M} \sum_{i=1}^n d(x, x_i)^2 \).  Each data point is mapped to the tangent space using  \(  u_i = x_i - \mu \). The covariance matrix, \( S = \frac{1}{n} \sum_{i=1}^n u_i u_i^T \), is constructed, and eigendecomposition yields principal directions \( v_k \) and variances are \( \lambda_k \), see \cite{flet} for the details.


To generalize these ideas to any data table, we define a Local Manifold Approximation, let \( X = \{x_1, x_2, \ldots, x_n\} \) be a dataset embedded in a high-dimensional space \( \mathbb{R}^p \). For each data point \( x_i \), UMAP defines a local neighborhood \( N_i \) consisting of its \( k \)-nearest neighbors under a metric \( d \). The \( k \)-nearest neighbors of \( x_i \) are defined as the set \( \mathcal{N}(x_i) \) such that:
\[
\mathcal{N}(x_i) = \left\{ x_j \in X \setminus \{x_i\} \; \middle|  \; d(x_i, x_j) \leq d(x_i, x_k), \, \forall x_k \notin \mathcal{N}(x_i) \right\},
\]
where:
\begin{itemize}
    \item \( d(x_i, x_j) \) is the distance between \( x_i \) and \( x_j \).
    \item \( k \) is the number of neighbors (a hyperparameter).
\end{itemize}

Thus, \( \mathcal{N}(x_i) \) is the set of the \( k \) closest points to \( x_i \) in \( \mathbb{R}^p \). The {\em Patch} \( \mathcal{P}_i \) is defined as a {\em fuzzy simplicial set} that approximates the local geometry of the manifold around \( x_i \). Formally:

\[
\mathcal{P}_i = \left\{ (x_i, x_j, w_{ij}) \, \middle| \, x_j \in N_i, \ w_{ij} > 0 \right\},
\]

\noindent where:
\begin{itemize}
    \item \( x_i \) is the anchor point.
    \item \( x_j \in \mathcal{N}_i \) are the \( k \)-nearest neighbors of \( x_i \).
    \item \( w_{ij} \) is the weight (or affinity) associated with the edge between \( x_i \) and \( x_j \), defined in the previews section.
\end{itemize}

The individual patches \( \mathcal{P}_i \) for all \( x_i \in X \) can be combined to form the global fuzzy simplicial set o graph \( G \), representing the entire manifold:

\[
G = \bigcup_{i=1}^n \mathcal{P}_i.
\]

Because the data resides on a Riemannian manifold, with local distances it is necessity to define something akin to a Riemannian correlation, requiring a Riemannian mean, and, more broadly, necessitating the development of {\em Riemannian Statistics}.  By leveraging the one-to-one relationship given by the Nerve Theorem in \ref{TN} and its corollaries, we define the vector subtraction the Riemannian correlation on manifold as follows.

\begin{definition} \label{def_rho}
Let $x_{\alpha}$ and $x_{\beta}$ rows of $X$, we define the subtraction induced by the UMAP algorithm as 
${x}_{\alpha} \ominus  {x}_{\beta}=\rho_{\alpha\beta}({x}_{\alpha} - {x}_{\beta}),$ where $
\rho_{\alpha\beta} = 1 - S_{\textsf{\tiny{UMAP}}}({x}_{\alpha},{x}_{\beta}).
$

\noindent Then, local distances generated by the UMAP algorithm can be defined as:

\[ d_{\textsf{\tiny{UMAP}}}(x_\alpha,x_\beta) = \|x_{\alpha} \ominus x_{\beta}\| = \sqrt{\langle x_{\alpha} \ominus x_{\beta}, x_{\alpha} \ominus x_{\beta} \rangle},
\]

\noindent where \( \langle \cdot, \cdot \rangle \) is an inner product on the vector space \( \mathbb{R}^p \). 
We also define the the addition induced by the UMAP algorithm as ${x}_{\alpha} = {x}_{\beta} \oplus  {x}_{\gamma}$ if ${x}_{\gamma} = x_{\alpha} \ominus x_{\beta}$.
\end{definition}

In the subsequent definition, we generalize Fréchet mean:

\begin{definition} \label{def_g}
Let $X \in M_{n \times p}$ the data table. We denote by $x_{1}, \ldots, x_{n} \in \mathbb{R}^p$ the rows of $X$ and by $y_{1}, \ldots, y_{p} \in \mathbb{R}^n$ the columns of $X$. Each vector $x_{i}$ can be also considered a point in the Riemannian manifold $M$ induced by the simplicial complex. The {\it Riemannian mean} is the minimizer of the sum-of-squared distances to the data:

$$
g=\arg \min_{x \in M} \sum_{i=1}^{n} d_{\textsf{\tiny{UMAP}}} \left(x, x_{i}\right)^{2}.
$$
\end{definition}

\begin{definition} 
We defined the variance-covariance matrix $ S \in M_{p \times p}$ of $X$ as
$ S = \frac{1}{n} \sum_{i=1}^n ({x}_{i} \ominus {g}) ({x}_{i} \ominus {g})^t$, where $({x}_{i} \ominus {g}) = \rho_{i\lambda} \left[\begin{array}{c}x_{i 1}-{g}_{1} \\ \vdots \\ x_{i p}-{g}_{p}\end{array}\right]$, note that ${g}$ must be equal to ${x}_{\lambda}$ for some $\lambda$. Then, we can define \textit{Riemannian correlation} between $y_i$ and $y_j$ columns of $X$, that are in $\mathbb{R}^n$, as follows
$R({y}_i,{y}_j)=\frac{S_{ij}}{\sqrt{S_{ii}S_{jj}}}$. 
\end{definition}

We can say that each patch \( \mathcal{P}_i \) induces a {\em Riemannian Submanifold}  $H_i$ with the local metric $d_{\textsf{\tiny{UMAP}}}$  generated by the UMAP algorithm, defined in previous paragraphs.
We are now ready to define {\em Riemannian Principal Component Analysis} (R-PCA) for any data $x_1, \dots, x_n \in \mathbb{R}^p$.  Our goal, analogous to PCA, is to find a sequence of subspaces $S_i$ de $ \mathbb{R}^p$ that maximize the projected variance of the data,  but, which also takes into account the local distances of each of the submanifods $H_i$ that we have in the data. We find a sequence of subspaces,  not a a sequence submanifolds,  because our total space is finite, then requiring the projections to be in there would greatly degrade the result, something that even happens in the PGA proposed in  \cite{flet}.

Taking into account that $G = \bigcup_{i=1}^n \mathcal{P}_i$ and the following properties of patches: 1) \( \mathcal{P}_i \) is a weighted graph over the local neighborhood \( \mathcal{N}_i \), where the weights \( w_{ij} \in [0, 1] \) define the strength of the connection between \( x_i \) and \( x_j \), 2) \( \mathcal{P}_i \) represents the local structure of the manifold around \( x_i \), capturing relationships with its neighbors, 3) the weight \( w_{ij} \) can be interpreted as a probability or affinity that \( x_i \) and \( x_j \) are connected within the underlying manifold,  the {\em Riemannian Principal Component Analysis} (R-PCA) that respects the local metrics of the submanifolds $H_i$,  can be defined by the Algorithm 1.

\begin{algorithm} \label{alg1}
\caption{Riemannian Principal Component Analysis (R-PCA)}
\begin{algorithmic}[1]
\State \textbf{Input:} $x_1, \dots, x_n \in \mathbb{R}^p$ and the number $k$ of nearest neighbors in UMAP algorithm.
\State \textbf{Output:} Principal directions $v_s \in S_i$, variances $\lambda_s \in \mathbb{R}$. 
\State Generate symmetric UMAP graph $G$.
\State Compute the matrix of UMAP similarities $S_{\textsf{\tiny{UMAP}}}(x_i,x_j)$ with $i,j = 1, \ldots, n$.
\State Compute the matrix $P = \rho_{ij}$ for $i,j = 1, \ldots, n$, as in definition   \ref{def_rho}.
\State Compute the Riemannian distance matrix $D$,  using definition   \ref{def_rho}.
\State Compute the Riemannian mean $g$ using $D$,  as in definition   \ref{def_g}.
\State Compute the Riemannian variance-covariance matrix $ S \in M_{p \times p}$ as
$$ S = \frac{1}{n} \sum_{i=1}^n ({x}_{i} \ominus {g}) ({x}_{i} \ominus {g})^t.$$
\State Compute the Riemannian correlation matrix $R\in M_{p \times p}$ where $R_{ij}=\frac{S_{ij}}{\sqrt{S_{ii}S_{jj}}}$.
\State Extract eigenvectors and eigenvalues of $R$: $\{v_s, \lambda_s\}$.
\State Compute the Riemannian Principal Components.
\State For the correlation circle, the Riemannian correlations between the original variables and the principal components are calculated.
\end{algorithmic}
\end{algorithm}

In practice, in algorithm 1 it is recommended to use at least $k = \left\lfloor \frac{n}{c} \right\rfloor $ (the whole part) where $c$ is the number of clusters that the data table is suspected to have, or the number of clusters that you want to study, that is, $k$ is at least the average number of individuals that each cluster has.

\section{Applications with simulated and real data}

\subsection{Description of the simulated data in \texttt{Data10D.csv}}

The \texttt{Data10D.csv} file contains data structured with 2900 rows and 10 variables,  such that the first two columns determine five clusters within the dataset. These clusters are explicitly identified in the \texttt{cluster} column. Additionally, the file includes eight extra columns that were generated following a specific process to ensure they do not alter the cluster assignments.

The purpose of this process was to extend the original dataset by adding new variables that provide additional information without modifying the existing classifications. Eight additional columns, named \texttt{var1} to \texttt{var8}, were created with values generated independently using a standard normal distribution \(N(0,1)\). This means that each new variable has a mean of 0 and a standard deviation of 1. These variables were generated randomly and independently of the first two columns and the clusters defined in the \texttt{cluster} column.

The process of generating these new variables began by determining the total number of rows in the original dataset. For each row, eight random values were generated using NumPy's \texttt{np.random.normal(0,1,n)} function, where \(n\) is the number of rows in the dataset. Each generated value was stored in one of the new columns, ensuring that all values followed the same distribution and were uncorrelated with one another or with the original variables.

The final result is a file that preserves the structure and cluster assignments of the original data while being enriched with eight new independent variables. The preservation of the \texttt{cluster} column ensures that the groupings defined by the first two columns remain intact, while the new variables provide additional information that can enhance analysis.

\begin{figure}[h]
\sidecaption
\includegraphics[scale=.40]{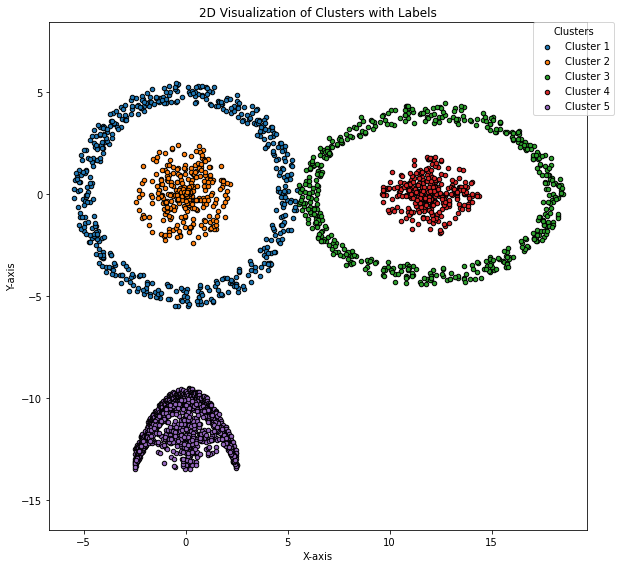}
\caption{Plotting the data using only the first two variables.}
\label{fig:Apl1}       
\end{figure}

The data in the file \texttt{Data10D.csv} corresponds to the 2D plot in Figure \ref{fig:Apl1}, where the first two columns (\texttt{x} and \texttt{y}) define the positions of the points and generate the clustering structure observed in the graph. These columns are responsible for the distinct shapes of the clusters, such as Cluster 1 and Cluster 2 are concentric, where Cluster 2 is nested within Cluster 1, forming a circle-within-a-circle structure.  Cluster 3 and Cluster 4 exhibit a similar nested structure but are located to the right of the first group, with Cluster 3 inside Cluster 4. Cluster 5 is located at the bottom of the graph, forming a distinct parabolic shape that is separated from the others. The \texttt{cluster} column in the file assigns each point to one of the five clusters, which is reflected in the graph through color coding: each cluster is represented by a unique color (e.g., blue for Cluster 1, orange for Cluster 2). While the additional eight variables (\texttt{var1} to \texttt{var8}) in the file were generated independently following a normal distribution \(N(0,1)\), they are not used in the 2D plot and do not influence the visualization. The plot accurately represents the clustering structure defined by the first two columns and the \texttt{cluster} column in the file, validating the data generation process and demonstrating the clear separability and arrangement of the clusters in the dataset.

\begin{figure}[h]
\sidecaption
\includegraphics[scale=.30]{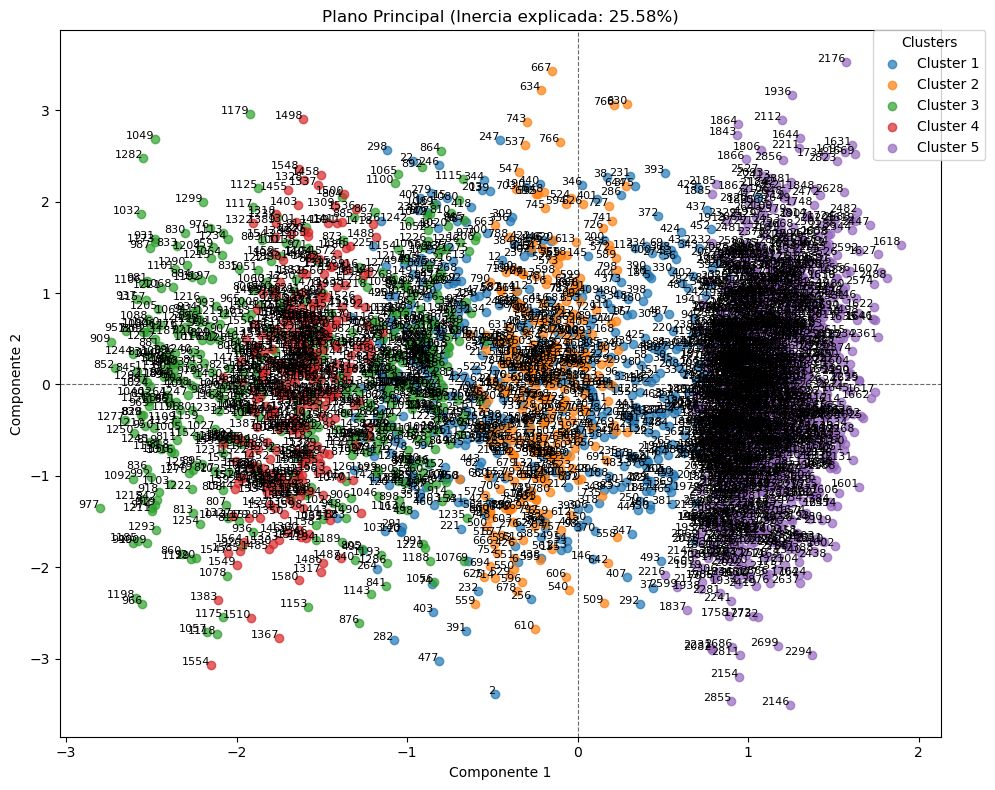}
\caption{PCA Plane of  \texttt{Data10D.csv}.}
\label{fig:Apl2}       
\end{figure}

\begin{figure}[h]
\sidecaption
\includegraphics[scale=.30]{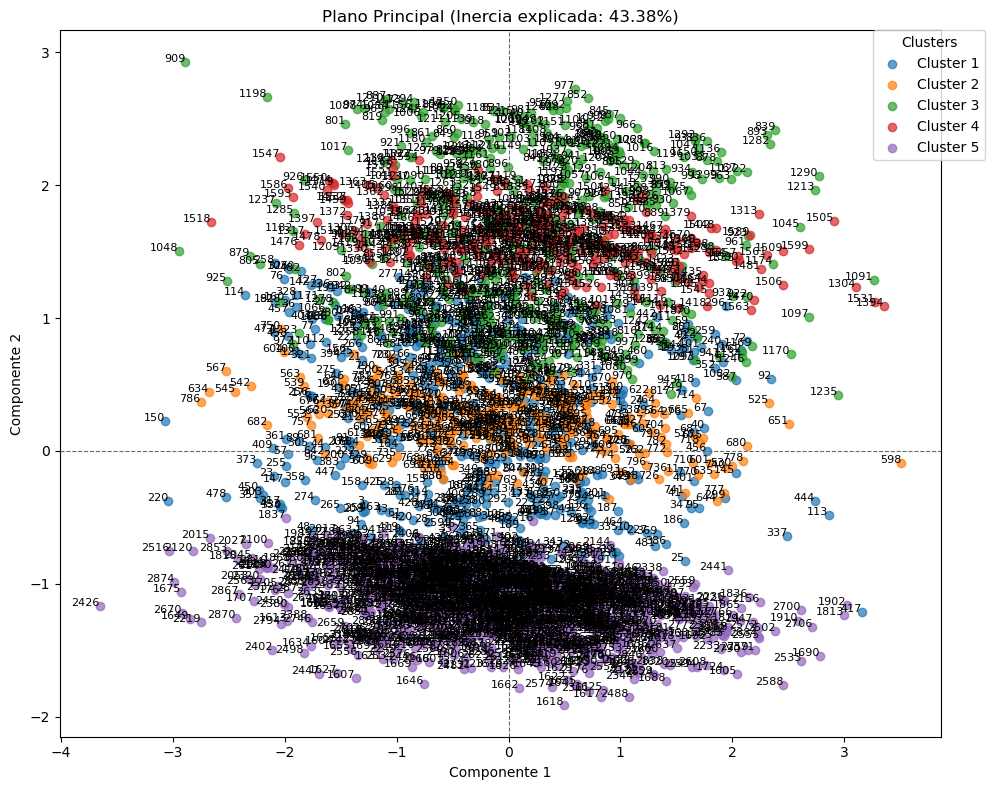}
\caption{R-PCA Plane of  \texttt{Data10D.csv}.}
\label{fig:Apl4}       
\end{figure}

\begin{figure}[h]
\sidecaption
\includegraphics[scale=.30]{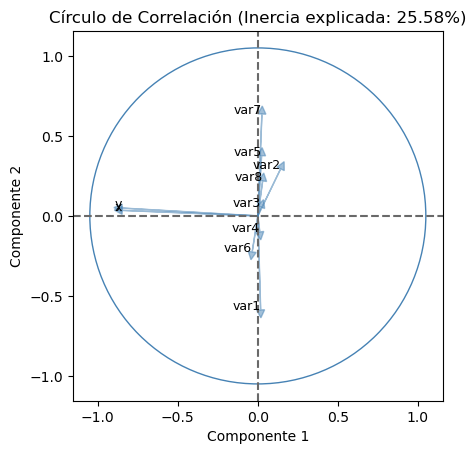}
\caption{PCA Correlations Circle of  \texttt{Data10D.csv}.}
\label{fig:Apl3}       
\end{figure}

\begin{figure}[h]
\sidecaption
\includegraphics[scale=.30]{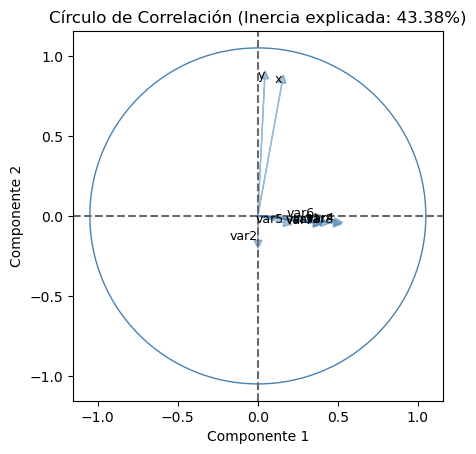}
\caption{R-PCA Correlations Circle of  \texttt{Data10D.csv}.}
\label{fig:Apl5}       
\end{figure}

The images \ref{fig:Apl2},  \ref{fig:Apl3},  \ref{fig:Apl4} and \ref{fig:Apl5} represent visualizations of the Principal Component Analysis (PCA) and the Riemannian Principal Component Analysis (R-PCA) applied to the dataset \texttt{Data10D.csv}. The plots \ref{fig:Apl2} and \ref{fig:Apl3}  correspond to the principal plane and the correlation circle of the PCA, respectively. Similarly,  figures \ref{fig:Apl4} and \ref{fig:Apl5}  represent the principal plane and the correlation circle of the R-PCA.  Here we are computing R-PCA with $k = \left\lfloor \frac{2900}{5} \right\rfloor = 580$.

The principal plane of the R-PCA in Figure  \ref{fig:Apl4} preserves the structure of the original five clusters in \texttt{Data10D.csv} more effectively than the standard PCA. The nested circular clusters (Clusters 1, 2, 3, and 4) and the parabolic cluster (Cluster 5) are clearly separated and retain their geometric shapes. In contrast, the principal plane of the PCA  in Figure  \ref{fig:Apl2} shows more diffuse clusters, with less distinct shapes.  Especially Cluster 5 (orange) which is a circle inside Cluster 1 (blue), is better captured in the R-PCA than the PCA showing the orange points mostly enclosed in the blue points. Similarly it happens between clusters 3 and 4 in green and red respectively.

The correlation circle of the R-PCA  in Figure \ref{fig:Apl5} provides a better interpretation of the relationships among the original variables, particularly highlighting the correlations for \texttt{x} and \texttt{y}. These variables are more prominently and accurately represented in the R-PCA correlation circle. In contrast, the PCA correlation circle  in Figure \ref{fig:Apl3} captures some variable relationships but is less effective at emphasizing these critical correlations.

The R-PCA achieves an explained variance (inertia) of \textbf{43.38\%}, which is significantly higher than the \textbf{25.58\%} explained by the PCA. This indicates that the R-PCA captures a larger proportion of the variability in the dataset, making it a more effective tool for dimensionality reduction in this context.

\subsection{Application to the real data set Olivetti Faces}

\begin{figure}[h]
\sidecaption
\includegraphics[scale=.30]{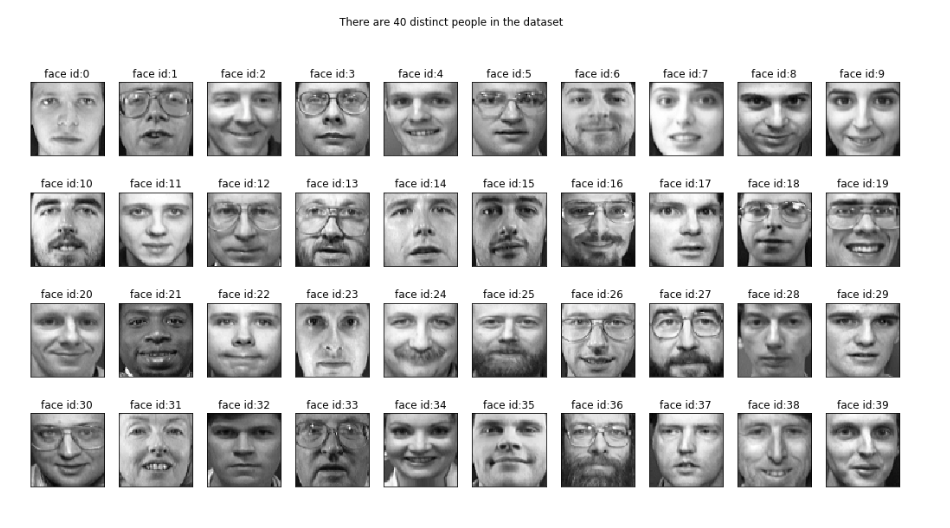}
\caption{Olivetti Faces data set.}
\label{fig:faces}       
\end{figure}

The \textit{Olivetti Faces} dataset is a widely used benchmark in the fields of pattern recognition, principal component analysis (PCA), and dimensionality reduction. Collected by the AT\&T Laboratories in Cambridge, the dataset contains \textit{400 grayscale images of 40 different individuals}, with \textit{10 distinct images per person} that vary in facial expression, detail, and lighting conditions. Each image has a resolution of \textit{64$\times$64 pixels}, resulting in \textit{4096 features per observation} when flattened into a one-dimensional vector. This dataset is commonly used to evaluate unsupervised learning algorithms such as PCA, $t$-SNE, and UMAP, as well as in facial recognition tasks. Its availability and well-structured nature make it a key tool for exploring multivariate statistical techniques on high-dimensional visual data. The Olivetti Faces is shown in Figure \ref{fig:faces},  we can see \cite{samaria} for more details.

\begin{figure}[h]
\sidecaption
\includegraphics[scale=.30]{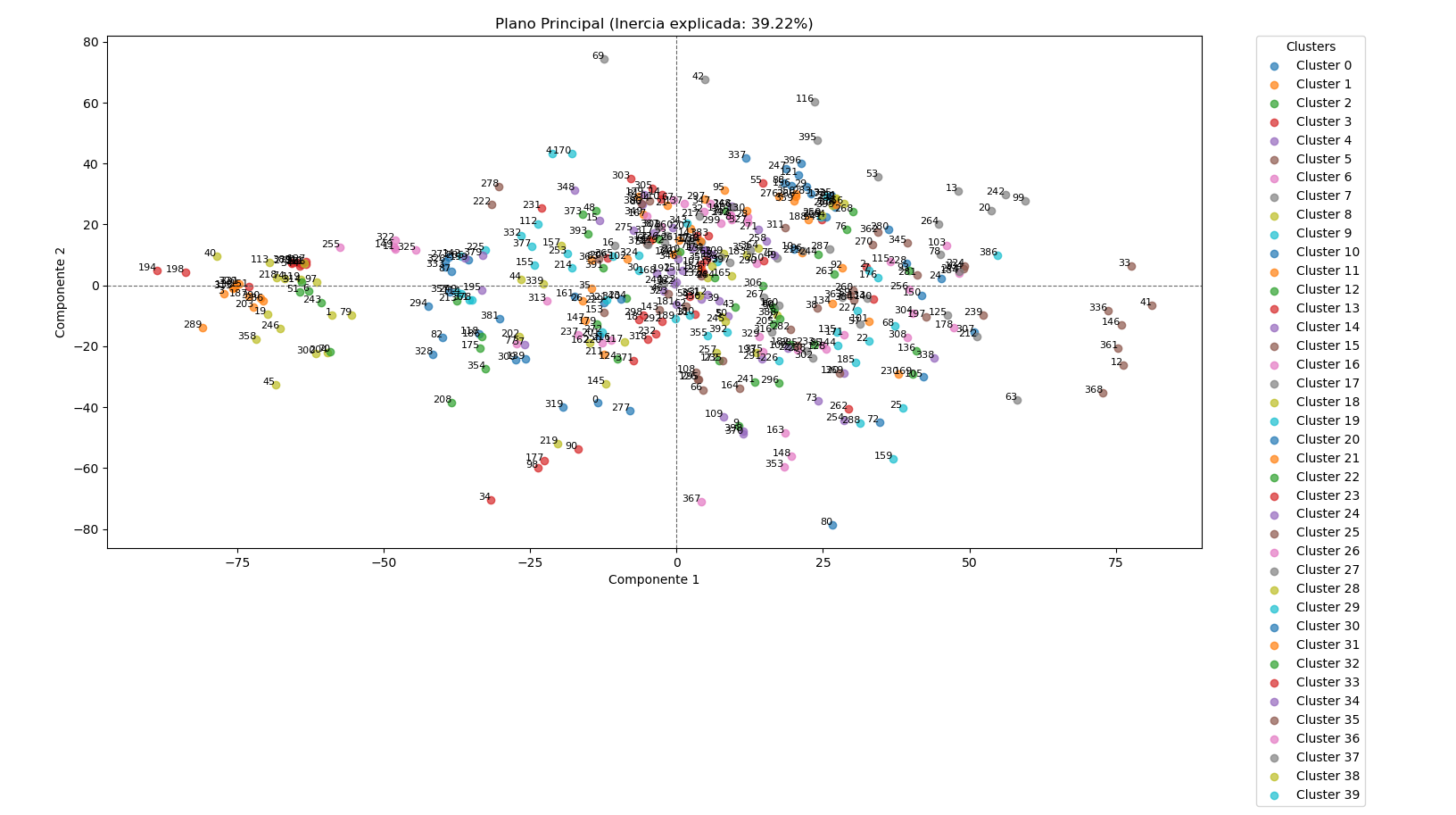}
\caption{Classic PCA in Olivetti Faces data set.}
\label{fig:PCAF}       
\end{figure}

\begin{figure}[h]
\sidecaption
\includegraphics[scale=.30]{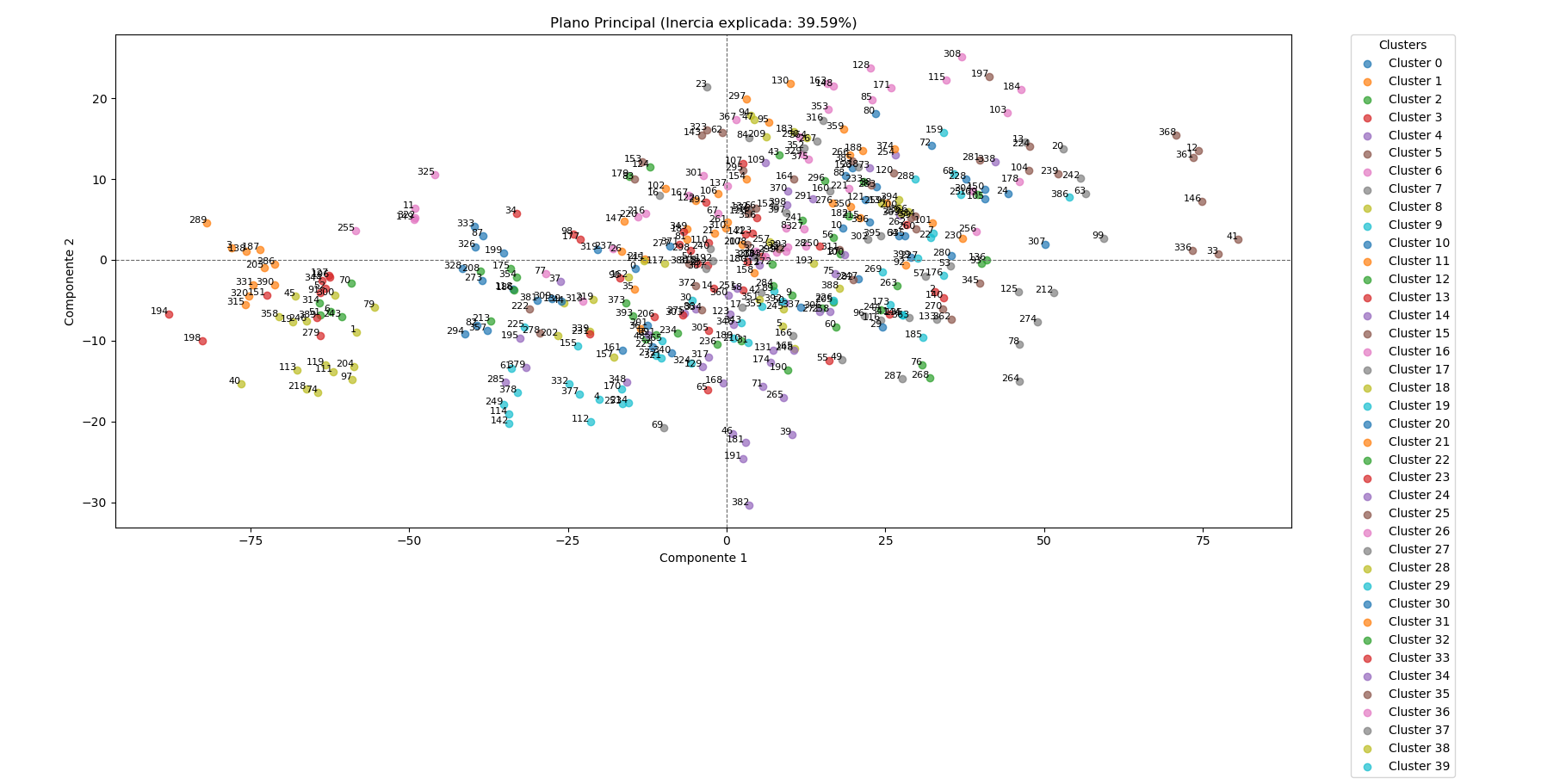}
\caption{R-PCA in Olivetti Faces data set.}
\label{fig:RPCAF}       
\end{figure}

When comparing the results of classical PCA,  in Figure \ref{fig:PCAF},   with the Riemannian PCA, in Figure \ref{fig:RPCAF}, differences arise in both variance explanation and cluster structure. First, the explained inertia in the first principal plane is higher in the R-PCA (39.59\%) than in the classical PCA (39.22\%), indicating that the Riemannian projection captures more information from the original data structure.

Moreover, the 40 clusters appear more clearly and homogeneously separated in the R-PCA projection, suggesting that this method is more effective in preserving inter-individual differences in the lower-dimensional space. In contrast, the classical PCA shows a higher degree of overlap between groups, making them harder to distinguish visually.

These results emphasize the benefit of extending linear techniques such as PCA to the context of Riemannian geometry, particularly for tasks like facial image analysis, where the underlying data space may be inherently non-Euclidean.

\section{Conclusions and Future Work}
\label{sec:conc}

In conclusion, the R-PCA outperforms the standard PCA in preserving the geometric structure of the clusters, such as the nested circles and the parabolic cluster, in the principal plane. It also provides a more meaningful interpretation of variable correlations in the correlation circle. The higher explained variance of the R-PCA further demonstrates its superiority in capturing the underlying structure of the data.

In this paper, we successfully extend the ideas proposed by \textit{Pennec et al.} in \cite{pennec}, broadening the scope to compute Riemannian statistical indices and Riemannian data analysis models to any data table. Unlike previous approaches, our methodology is not restricted to data with an intrinsic Riemannian manifold structure. This advancement opens up a new field of research, where diverse methods like regression, $k$-means, and more, can be generalized for broader applicability.

The \textit{Riemannian STATS Package}, available on GitHub \cite{rodriguez},  is a Python package developed to extend classical multivariate statistical methods to the Riemannian framework. It enables the transformation of a Euclidean data table into its Riemannian counterpart by applying UMAP algorithm. This tool is particularly valuable for analyzing datasets where the linear assumptions of traditional methods do not hold, allowing for more accurate representation and interpretation of the underlying data structure. Detailed information and examples about this package can be found on the website \url{https://riemannianstats.web.app/index.html}.

Currently, we are also working on a version of this package implemented in the  R programming language, which will offer similar functionality and integration with the R ecosystem for statistical computing and data analysis.

\end{document}